\documentclass[letterpaper, 10 pt, journal, twoside]{IEEEtran}

\usepackage{times}
\usepackage{epsfig}
\usepackage{graphicx}
\usepackage{amsmath}
\usepackage{amssymb}
\usepackage{array}
\usepackage{booktabs}
\usepackage{multirow}
\usepackage{rotating}
\usepackage{siunitx}
\usepackage{ulem}
\usepackage{cite}
\usepackage{subfigure}
\usepackage{hyperref}
\usepackage[export]{adjustbox}

\usepackage{color}

\newcommand{\change}[1]{{#1}}
\newcommand{\changeAmadeus}[1]{{#1}}
\newcommand{\qmarks}[1]{``#1"}

\begin{document}

\title{Augmenting Visual Place Recognition with Structural Cues}

\author{Amadeus Oertel, Titus Cieslewski, and Davide Scaramuzza%
\thanks{Manuscript received: February, 24th, 2020; Revised May, 27th, 2020; Accepted June, 19th, 2020.}%
\thanks{This paper was recommended for publication by Editor Cesar Cadena Lerma upon evaluation of the Associate Editor and Reviewers' comments.
This work was supported by the National Centre of Competence in Research Robotics (NCCR) through the Swiss National Science Foundation and the SNSF-ERC Starting Grant.}
\thanks{The authors are with the Robotics and Perception Group, Dept. of Informatics, University of Zurich, and Dept. of Neuroinformatics, University of Zurich and ETH Zurich, Switzerland---\url{http://rpg.ifi.uzh.ch}.}
\thanks{Digital Object Identifier (DOI): see top of this page.}
}

\markboth{IEEE Robotics and Automation Letters. Preprint Version. Accepted June, 2020}
{Oertel \MakeLowercase{\textit{et al.}}: Augmenting Visual Place Recognition with Structural Cues}

\maketitle

\begin{abstract}

In this paper, we propose to augment image-based place recognition with structural cues.
Specifically, these structural cues are obtained using structure-from-motion, such that no additional sensors are needed for place recognition.
This is achieved by augmenting the 2D convolutional neural network (CNN) typically used for image-based place recognition with a 3D CNN that takes as input a voxel grid derived from the structure-from-motion point cloud.
We evaluate different methods for fusing the 2D and 3D features and obtain best performance with global average pooling and simple concatenation.
\change{On the Oxford RobotCar dataset,} the resulting descriptor exhibits superior recognition performance compared to descriptors extracted from only one of the input modalities, including state-of-the-art image-based descriptors.
Especially at low descriptor dimensionalities, we outperform state-of-the-art descriptors by up to 90\%.
   
\end{abstract}


\IEEEpeerreviewmaketitle

\section{Introduction}

\IEEEPARstart{P}{lace} recognition is a key concept for localization and autonomous navigation, especially so in GPS-denied environments.
In particular, the ability to recognize a previously observed scene is a fundamental component in Simultaneous Localization and Mapping (SLAM).
Here, place recognition is used to provide loop closure candidates, which can be used to compensate for accumulated drift, thereby enabling globally consistent mapping and tracking.
Furthermore, place recognition can support the purpose of localization with respect to a pre-built map of the environment~\cite{Sarlin18corl}.  %
The de-facto standard approach involves casting its formulation as an image-retrieval problem \cite{Cummins08ijrr} in which a query image is matched to the most similar one in a database of images representing the visual map.
The 6 degree-of-freedom pose of the query camera frame can subsequently be inferred from the retrieved database image. 

\begin{figure}[t]
\begin{center}
   \includegraphics[width=1.0\linewidth]{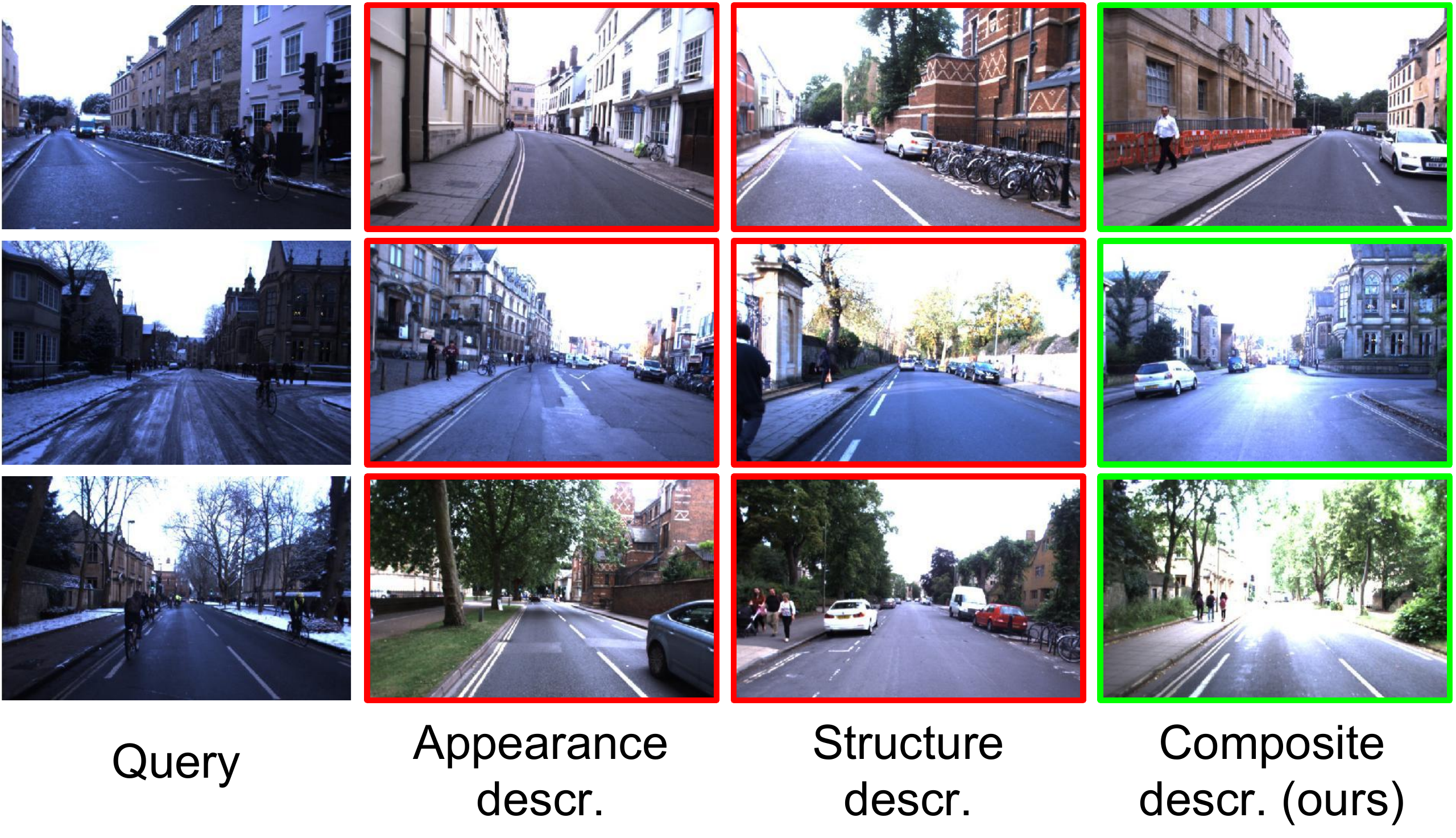}%
   \vspace{-3mm}
   \caption{To enhance visual place recognition, we propose a composite descriptor that combines both visual and structural cues -- obtained using structure from motion -- in a single representation. 
   This figure illustrates three queries and the retrieved image using different types of descriptors. 
   While descriptors that separately leverage either appearance or structure produce incorrect matches, each query is successfully matched using our composite descriptor despite changes in viewpoint and visual appearance.
   \change{See the supplementary material~\cite{Oertel20ralSupplementary} for further examples.}}
   \label{fig:retrieval_comparison}
\end{center}
\vspace{-9mm}
\end{figure}

Traditionally, image matching has been achieved by first extracting handcrafted sparse local feature descriptors \cite{Lowe04ijcv, Alahi12cvpr, Rublee11iccv}. 
Related research has focused on various ways to efficiently match such local features and to meaningfully aggregate them for generating matches between full images \cite{Sivic03iccv, Cummins08ijrr, Jegou10cvpr, Philbin07cvpr, Arandjelovic13cvpr, Jegou12pami, Lowry16tro}.
However, challenges arise when images representing the same location are captured under strong variations in appearance~\cite{Maddern16ijrr}.
These variations can be due to changes in illumination, weather conditions, and camera viewpoint.
Considering the ultimate goal of achieving large-scale long-term place recognition, the problem setting is further complicated by severe seasonal changes in appearance, structural scene modifications over time, for example due to roadworks, and perceptual aliasing in repetitive environments.

To cope with these challenges, recent works focus on replacing the handcrafted local features with convolutional neural networks (CNNs), which have proven useful in various object-level recognition tasks \cite{Krizhevsky12nips, Girshick14cvpr, Long15cvpr}.
CNNs are used to either improve the discriminability of \textit{learned} local 2D image features \cite{Loquercio17icra, Noh17cvpr} or directly learn powerful global image descriptors in an end-to-end manner \cite{Arandjelovic16cvpr, Gordo16eccv, Radenovic18pami}. 

However, due to the susceptibility of image-based approaches to variations in scene appearance, a different branch of research has developed local \cite{Cieslewski16icra, Ye17bmvc} and global \cite{Cramariuc18arxiv, Uy18cvpr} 3D structural descriptors to perform place recognition in 3D reconstructed maps.
Most of these structural descriptors are based on maps generated from LiDAR scanners or RGBD cameras. 
As an exception, \cite{Cieslewski16icra, Ye17bmvc} proposed a local descriptor matching approach in which learned descriptors are derived from the point cloud obtained by structure-from-motion.
These structural descriptors were compared to state-of-the-art appearance-based descriptors, with mixed results -- in some cases, structure outperforms appearance, while in other cases, appearance outperforms structure.
Inspired by these results, we show in this paper that place recognition performance can improve significantly by learning to incorporate both visual and 3D structural features, that contain information extracted over \textit{sequences} of images, into a unified location descriptor. 
To this end, we propose a simple yet effective deep CNN architecture that is trained end-to-end on the combination of both input modalities.
We demonstrate the superior matching and retrieval performance of the resulting descriptors compared to descriptors of the same dimensionality extracted from only one of the two input domains.
Specifically, the structural features are obtained from image sequences using structure from motion, such that no additional sensors are needed to deploy our method in practice.
We verify our method throughout several experiments by comparing its retrieval performance to state-of-the-art baselines including NetVLAD~\cite{Arandjelovic16cvpr}, DenseVLAD~\cite{Torii18pami}, SeqSLAM~\cite{Milford12icra}, and \change{Multi-Process Fusion~\cite{Hausler19ral}}.
To the best of our knowledge, we are the first to propose learned composite descriptors that incorporate both appearance and structure for the task of visual place recognition. 
\change{To summarize, we propose to derive place recognition descriptors from both appearance and structure and show that the resulting descriptor outperforms variants obtained from only one of the two modalities.
Several methods for fusing visual and structural features are evaluated, and we show that of the evaluated methods, the simplest one -- concatenation of globally pooled deep convolutional features -- results in best performance.}

\vspace{-2mm}
\section{Related Work}

Our related work can be broadly divided into methods that rely on matching of either images or structural segments.%
\medskip

\noindent \textbf{Place recognition in visual maps.} Visual place recognition is typically treated as an image retrieval problem \cite{Cummins08ijrr} where the task of recognizing places is solved by matching individual images of the same location.
Classically, this has been achieved by extracting \textit{handcrafted} sparse local feature descriptors, such as SIFT \cite{Lowe04ijcv}, ORB \cite{Rublee11iccv}, or FREAK \cite{Alahi12cvpr}, at salient regions of the images.
\change{Subsequently, such local descriptors} are typically combined to form a global descriptor for each image using aggregation methods such as bag-of-visual-words \cite{Sivic03iccv, Jegou10cvpr, Philbin07cvpr}, VLAD \cite{Arandjelovic13cvpr}, or Fisher Vectors (FV) \cite{Jegou12pami}\change{, which allow for direct matching of the represented images}.
Alternatively, matches between local features can be \change{accumulated} to a match between images using nearest-neighbour voting schemes \cite{Jegou08eccv, Lynen14threedv}.
State-of-the-art performance has been \change{demonstrated} using VLAD with descriptors calculated at every pixel, rather than at sparse locations~\cite{Torii15cvpr}.
See \cite{Lowry16tro} for a survey of handcrafted visual place recognition methods.

Early retrieval methods that use convolutional neural networks (CNNs)  have employed off-the-shelf networks, usually pre-trained on ImageNet for image classification, as black box feature extractors \cite{Razavian14cvprw, Babenko14eccv}. 
While able to outperform retrieval based on traditional global representations, these approaches do not reach the performance of previous methods based on local descriptor matching.
Therefore, subsequent works have proposed hybrid approaches based on conventional aggregation of local CNN features including FV \cite{Perronnin15cvpr} and VLAD \cite{Gong14eccv, Paulin15iccv}.
More recently, architectures have been proposed that allow the training of full image descriptors in an end-to-end fashion.
These approaches typically interpret the output of a CNN as densely \change{extracted} descriptors that are \change{subsequently} aggregated with a differentiable pooling method like NetVLAD~\cite{Arandjelovic16cvpr} or Generalized-Mean Pooling~\cite{Radenovic18pami}, resulting in state-of-the-art performance on various benchmarks.
\medskip

\noindent \textbf{Place recognition in structural maps.}
Despite advancements in CNN-based retrieval, the proneness to strong changes in visual appearance of a scene constitutes a major disadvantage of place recognition systems using single images.
As an alternative, using 3D structure can offer benefits in terms of robustness to challenging environmental scene alterations and changes in illumination.
As in image representations, early works propose \textit{handcrafted} local descriptors \cite{Rusu08iros, Tombari10eccv, He16iros, Dube17icra, Bosse13icra}, while more recent works leverage neural networks.
Among the latter, two approaches to extracting structural features from point clouds have emerged:
in the first approach, 3D points are aggregated into cells of a 3D grid, and features are learned using 3D convolutions \cite{Cramariuc18arxiv,Zeng17cvpr}.
In the second approach, features are learned directly from the point cloud using multi-layer perceptron responses to point locations \cite{Qi17cvpr,Uy18cvpr}.
All aforementioned works use dense point clouds, as extracted from specific sensors such as LiDAR scanners or RGBD cameras.
{In contrast}, we focus on situations where only visual information is available.

It has \change{also} been shown that place recognition \change{is possible based on} the sparser structural data \change{generated by} vision-based structure-from-motion (SfM) algorithms \cite{Cieslewski16icra, Ye17bmvc}.
A \change{key} difference to the denser LiDAR or RGBD data\change{, which is} particularly relevant \change{to extracting} handcrafted descriptors \cite{Cieslewski16icra}, is that surface normals cannot be \change{computed} reliably.
\change{Therefore, like} \cite{Zeng17cvpr}, the CNN-based \change{approach} of \cite{Ye17bmvc} uses features learned from a 3D grid \change{representation}.
Compared to \change{visual} place recognition, they report mixed results -- in some cases, structure outperforms appearance, \change{while} in other cases appearance outperforms structure.

Hence, to leverage useful features from both \change{modalities}, we propose \change{fusing them to form} a unified descriptor.
{This idea} has very recently been applied {to} object detection \cite{Liang18eccv} and bounding box regression \cite{Xu18cvpr}.
To the best of our knowledge, we are the first to apply \change{this concept} to place recognition.
Note that structure-from-motion \change{represents} information extracted over a sequence of images.
In contrast, approaches like SeqSLAM~\cite{Milford12icra} \change{and more recently Multi-Process Fusion (MPF) \cite{Hausler19ral}} use image sequences directly.
\change{Specifically, MPF presents an alternative approach by fusing information from multiple image processing methods including histogram of gradients and CNN features, showing significant improvements over SeqSLAM and NetVLAD.}
\begin{figure*}
	\centering
	\subfigure[]{%
		\label{fig:architecture}%
		\includegraphics[width=0.65\linewidth,trim={0 6mm 0 6mm},clip]{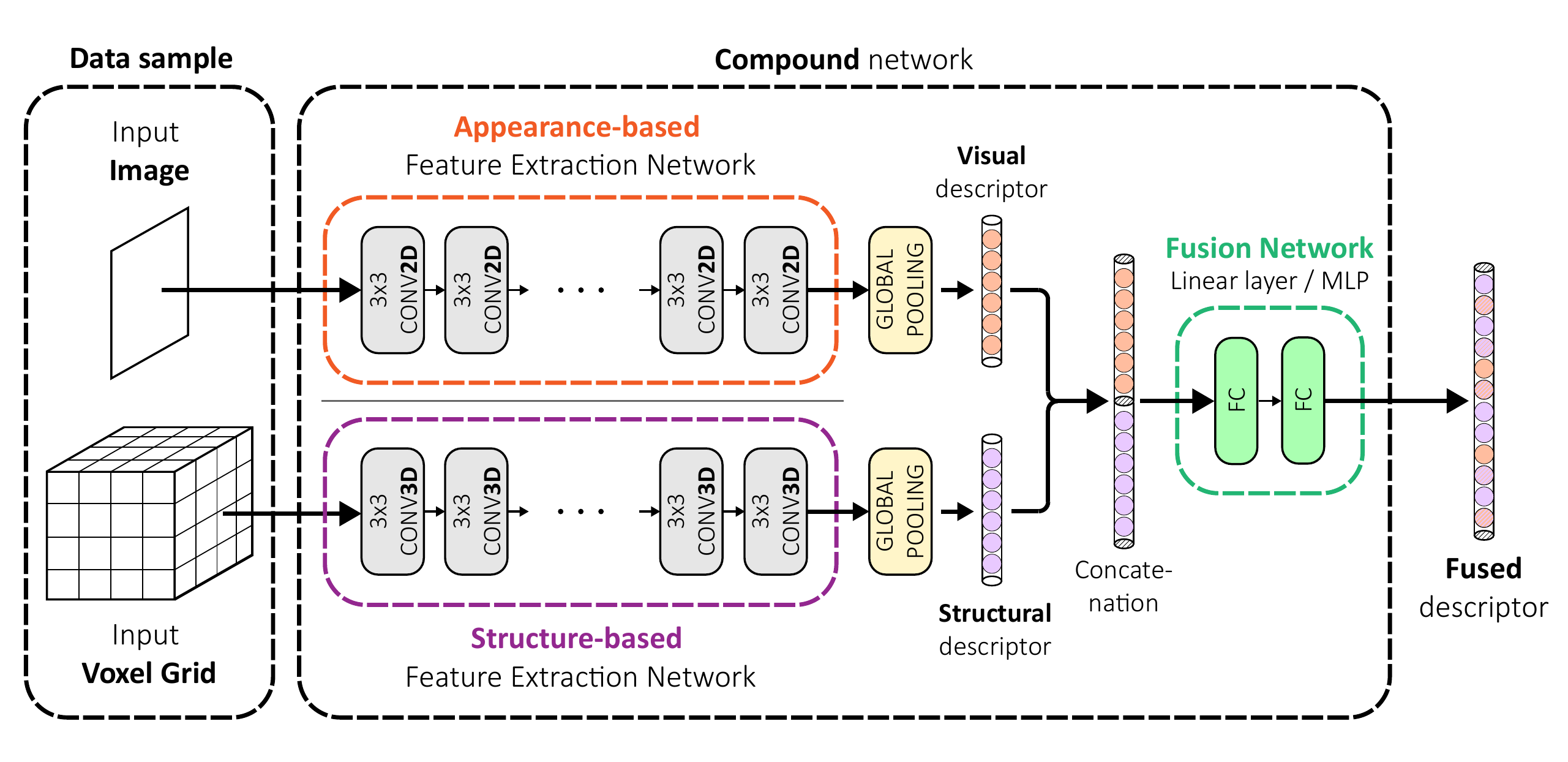}}%
	\hfill
	\hspace{0.5mm}
	\subfigure[]{%
		\label{fig:hard_mined}%
		\includegraphics[width=0.31\linewidth, height=4.8cm]{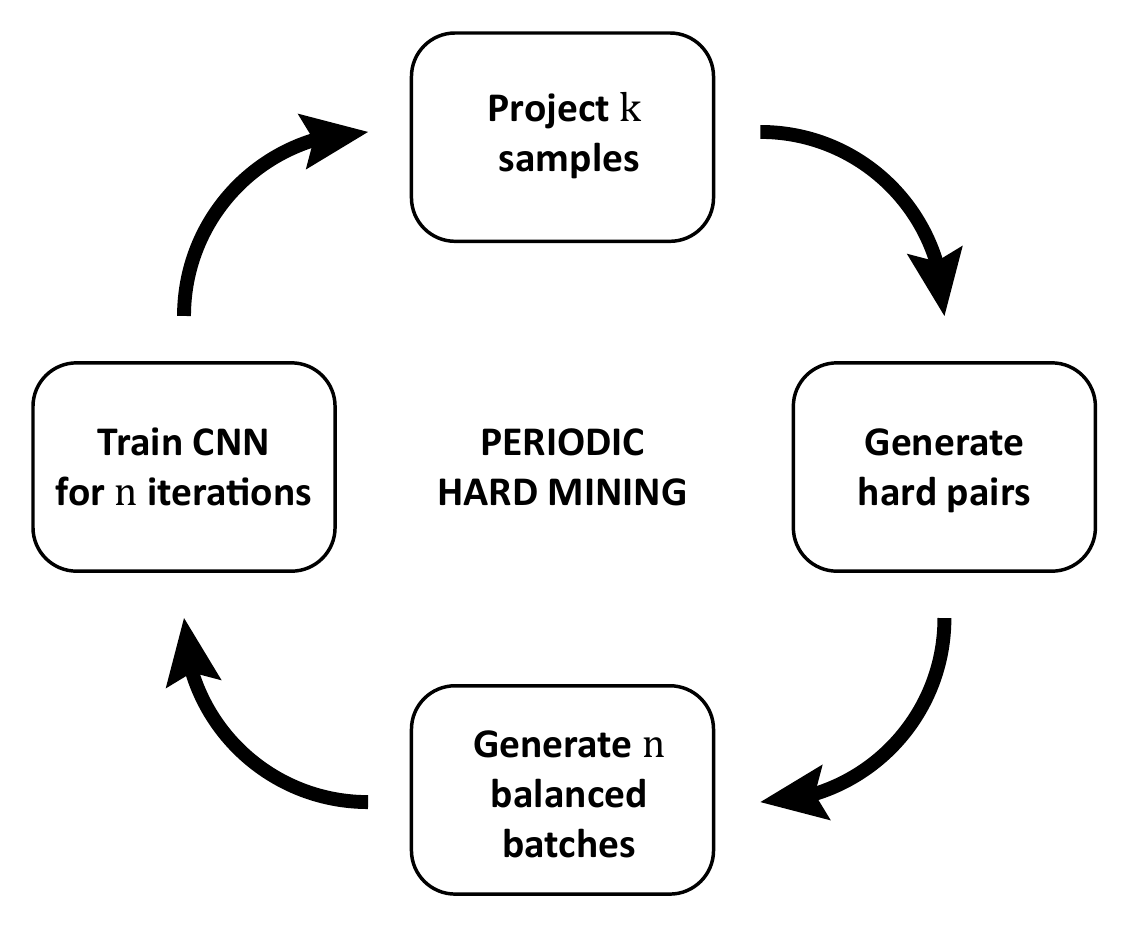}}%
	\vspace{-1.5mm}
	\caption{\change{\subref{fig:architecture} Overview of our proposed network architecture consisting of three building blocks including two separate CNN-based feature extraction networks followed by an optional fusion network. \subref{fig:hard_mined} Schematic of our hard mining policy which is a modified version of that presented in \cite{Loquercio17icra}.}}
	\vspace{-7mm}
\end{figure*}

\section{Methodology}
In this section, we describe our general network architecture as well as our training methodology.
In our place recognition system, an observation $i$ of a place is composed of an image $I_i$ and a voxel grid $G_i$ encoding the corresponding local 3D structure.
\change{We describe} the extraction of \change{$G_i$} from an image sequence in Section \ref{sec:voxel_grids}.
\change{
While it is possible to use alternative input representations for feeding 3D structure into a neural network, such as in PointNet~\cite{Qi17cvpr} or PointNet++~\cite{Qi17nips}, we use a grid representation based on the methodologies of~\cite{Ye17bmvc, Cramariuc18arxiv, Dube17icra}.}
Our goal is to learn the function $f_{\theta}$, parametrized with a set of parameters $\theta$, that maps $I_i$ and $G_i$ to a descriptor vector $f_{\theta}(I_i, G_i)$, such that the distance $d(f_{\theta}(I_i, G_i), f_{\theta}(I_j, G_j))$ is small if observations $i$ and $j$ are of the same place, and large otherwise.
While any distance metric can be used as $d$, we have empirically found the $L_1$ distance to work best in our experiments.
The goal is to train $f_{\theta}$ such that this property is maintained under strong viewpoint and natural appearance changes between different observations.

\subsection{Architectural Details}
\label{sec:net_details}
The overall structure of our proposed \textit{compound} network is summarized in Fig. \ref{fig:architecture}.
The three main components comprise two fully convolutional networks \cite{Shelhamer17pami} dedicated to visual and structural \textit{feature extraction}, each producing a descriptor specific to its modality, and an optional \textit{feature fusion} network that combines both sets of features into a composite descriptor.
We investigate different approaches to combine the intermediary descriptors ranging from simple concatenation to propagating the concatenated descriptors through a multi-layer perceptron. \medskip

\noindent \textbf{Feature extraction.}
Features are separately extracted for the image $I_i$ and the voxel grid $G_i$, respectively.
The visual feature extraction network is composed of \change{twelve} $~ 3 \times 3$ convolutional layers with stride $1$ while every other layer starting with the second one is followed by non-overlapping $2 \times 2$ max-pooling with stride $2$. 
The first \change{six} convolutional layers use $64$ output channels, which is doubled to $128$ for the remaining layers. 
The structural feature extraction network is composed analogously to its visual counterpart with the exception of using \change{nine} $~ 3 \times 3 \times 3$ convolutional layers interspersed with 3D \textit{average} pooling operations. 
For computational efficiency, the first two layers are fixed at $32$ output channels.
The \change{subsequent four} layers {use} $64$ and the \change{remaining three layers use} $128$ channels, respectively.
Both networks apply a Rectified Linear Unit (ReLU) non-linearity to the output of every convolutional layer.
Note that \change{these are the configurations describing our best performing compound model and deviate} slightly for some of the feature extraction networks evaluated separately in Section \ref{sec:discussion}. \change{Refer to our supplemental material~\cite{Oertel20ralSupplementary} for further information.} %
\medskip

\noindent \textbf{Global average pooling.}
As shown in Fig. \ref{fig:architecture}, we apply global average pooling~\cite{Radenovic18pami} to the output of each extraction network's final convolutional layer, resulting in $c_f$-dimensional visual and structural descriptors $g_A$ and $g_S$. 
$c_f=128$ for the standard configuration described above.
Compared to using fully-connected (FC) layers, global average pooling results in robustness to spatial translations, reduces proneness to overfitting, and allows processing of arbitrarily-shaped inputs, which is particularly interesting for changes in camera resolution across applications \cite{Lin14iclr}.
\change{We use no other form of normalization besides the division by the image resolution involved in average pooling.}
\medskip

\noindent \textbf{Feature fusion.}
We evaluate and compare four different methods to obtain a unified representation from intermediary visual and structural feature vectors.
The first one is a \textit{simple concatenation}, such that the final descriptor is given by
\begin{equation}\label{eq:concat}
	f_{\theta}(I_i, G_i) = \big[ g_A( I_i ) \; \; g_S( G_i ) \big] \,.
\end{equation}
The second approach is a \textit{weighted concatenation} using two learnable scalar weights $w_A$ and $w_S$ to rescale each of the intermediary feature vectors prior to concatenation,
\begin{equation}\label{eq:weight_concat}
	f_{\theta}(I_i, G_i) = \big[ w_A \cdot g_A( I_i ) \; \; w_S \cdot g_S( G_i ) \big] \,.
\end{equation}
To allow for more flexibility in learning correlations across feature domains, the third version of our compound architecture trains a $dim_f \times 2 c_f$ \textit{linear projection} $W_c$ of the concatenation, as defined by
\begin{equation}\label{eq:linear}
	f_{\theta}(I_i, G_i) = W_c \big[ g_A( I_i ) \; \; g_S( G_i ) \big] \,.
\end{equation}
Finally, we use a \textit{multi-layer perceptron} $g_{mlp}$ to investigate learning a non-linear mapping between concatenated features and final composite descriptor,
\begin{equation}\label{eq:linear}
	f_{\theta}(I_i, G_i) = g_{mlp} \left( \big[ g_A( I_i ) \; \; g_S( G_i ) \big] \right) \,.
\end{equation}
The standard configuration employs two FC layers with 256 units each, followed by ReLU activation.

\subsection{Training Methodology}
\label{sec:train_method}
\change{Our network is trained using a margin-based loss on training samples which are pairs of inputs labeled with whether they should result in a positive or a negative match.
While some place recognition methods are trained on samples from the database sequence used during deployment, we want our network to learn invariance to appearance conditions, so our training set needs to represent such changes in appearance.
Furthermore, our goal is to obtain a system which can be deployed in different environments without the need for retraining, so we use strictly disjoint sets of training and evaluation samples.}
In this section, we describe our choice of loss function and the batch sampling strategy we have used for training. 
Training has been done using stochastic gradient descent on a single Graphics Processing Unit (GPU) -- an Nvidia Titan Xp or an Nvidia RTX 2080 Ti.
\vspace{1mm}

\noindent \textbf{Margin-based loss.}
We associate each observation $i$ with a data sample $\mathbf{X}_i = \left(I_i, G_i \right)$ consisting of an image $I_i$ and corresponding voxel grid $G_i$.
The network learns to distinguish matching and non-matching locations by optimizing a \textit{margin-based loss} \cite{Wu17iccv} that aims at minimizing the $L_1$ distance $d(\mathbf{X}_i, \mathbf{X}_j)$ between a pair of mapped samples $\mathbf{X}_i, \mathbf{X}_j$ corresponding to matching observations while maximizing that of non-matching observations.
With the hinge loss denoted as $\ell\left( x \right)=max\left(x, 0\right)$, the margin-based loss function is given by
\begin{equation}\label{eq:margin_loss}
	\mathcal{L}\big(y_{ij}, \mathbf{X}_i, \mathbf{X}_j\big) = \ell \Big(\alpha + y_{ij} \cdot \big(d(\mathbf{X}_i, \mathbf{X}_j) - m \big) \Big) \,, 
\end{equation}
where $y_{ij} \in \{-1, 1\}$ denotes the ground truth of whether both descriptors should match and the parameters $\alpha$ and $m$ are such that the distance between matching descriptors is nudged below $m - \alpha$ while the distance between non-matching descriptors is nudged above $m + \alpha$. %
Compared to the more commonly known contrastive loss \cite{Hadsell06cvpr}, a margin-based loss allows matching samples to be within a certain distance of each other rather than enforcing them to be as close as possible.
A very popular alternative is given by the \textit{triplet loss} \cite{Schulz03nips} which has been widely applied to different problem settings \cite{Schroff15cvpr, Loquercio17icra}.
However, \cite{Wu17iccv} highlights the importance of sampling for deep embedding learning and shows that a simple margin-based loss is capable of outperforming other losses including contrastive and triplet loss on the task of image retrieval and a range of related tasks.
Indeed, preliminary experiments using margin-based versus triplet loss resulted in faster convergence and better discriminative power, especially in conjunction with the hard mining strategy described next.
\vspace{1mm}

\noindent \textbf{Batch sampling strategy.}
Uniform random sampling from all possible training pairs rapidly yields an increasing fraction of \qmarks{easy} pairs that result in a loss of zero.
To avoid significantly prolonged convergence times and inferior performance of the converged models, we adopt the concept of hard mining previously exploited in similar learning-based settings \cite{Simoserra15iccv, Ye17bmvc, Loquercio17icra}.
In hard mining, training is focused on samples with high loss.
\qmarks{Hard} samples can be determined using forward passes, which are computationally cheaper than training passes.
In our particular pair-based training setup, we exploit the fact that the loss of $x^2$ pairs can be evaluated using only $2x$ descriptor forward passes.
The set of hard samples technically changes with every iteration, but it would be prohibitive to look for the hardest pair of the entire training dataset at every iteration.
Consequently, we rely on a randomized hard mining strategy, where the $n$ hardest samples are selected from $k^2$ random pairs every $n$ training iterations, see Fig. \ref{fig:hard_mined}.
$k$ and $n$ are iteratively adapted \change{as training progresses}:
$k$ is increased over time, and if there are too many samples with zero loss, $n$ is decreased.
To avoid focus on outliers and to account for the natural imbalance between true negatives and true positives, we use a balanced batch composition equally divided into hard pairs, randomly selected positive pairs, and randomly selected negative pairs.
We use the largest batch size divisible by three that fits GPU memory - $12$ when training the compound network and $24$ when training the feature extraction networks individually.
\subsection{Voxel Grid Representation}
\label{sec:voxel_grids}
The goal of our setup is to \change{rely on vision sensors only}, and so we use 3D structural information extracted from the input image sequences.
The reconstructed 3D point segments are discretized into regular voxel grids to represent local structural information.
In principle, any kind of Visual Odometry~\cite{Forster17troSVO} or SLAM~\cite{MurArtal17tro, Engel14eccv} framework may be used for 3D point cloud generation from image sequences.
However, we are interested in exploiting the rich structural information provided by semi-dense reconstructions following the promising results reported in \cite{Ye17bmvc}.
To this end, we use a variant of the publicly available \textit{Direct Sparse Odometry} (DSO) framework \cite{Engel17pami} extended to perform pose tracking and mapping using stereo cameras \cite{Wu18git}.
As a direct method, DSO is able to track and triangulate all image points that exhibit intensity gradients, including edges. 

Given the 3D reconstruction of an image sequence, we generate one voxel grid per image.
A point cloud submap is extracted from a rectangular box centered at the camera pose at which the image was taken.
We assume each submap to be aligned with both the $z$-axis of the world frame which can be achieved using an Inertial Measurement Unit (IMU), and with the yaw orientation of the corresponding camera pose.
The size of the box is fixed in our method and needs to be adapted according to the environment in which it is used.
A submap contains all points observed by DSO over the set of $N$ preceding keyframes (ending with the frame associated with this submap) that are located within the box boundaries.
Next, the submap is discretized into a regular voxel grid.
We evaluate three different methods to populate the grids:
with \textit{binary occupancy} (bo), a voxel with any 3D points located inside is assigned a value of 1 and 0 otherwise.
With \textit{point count} (ptc), each voxel value is equal to the number of 3D points within that voxel.
With \textit{soft occupancy} (so), each 3D point receives a weight equal to 1.0 which is distributed among the eight nearest voxel centers using tri-linear interpolation. 
We compare the performance achieved when using each of these representations in Section \ref{sec:grid_type}.

\section{Experiments}
In this section, we describe the dataset and our evaluation methodology, and provide quantitative and qualitative results to validate our approach.

\setlength{\heavyrulewidth}{1.5pt}
\setlength{\abovetopsep}{6pt}
\begin{table}
\footnotesize
\centering
\caption{Number of images per condition and season part of our training and testing sets, which are selected from the original sequences with timestamps listed below.}
\vspace{-4.5mm}
\begin{tabular}{p{0.038 \textwidth} p{0.08 \textwidth}}
\toprule
\multirow{2}{*}{\scriptsize{\textbf{Train}}} & \multicolumn{1}{l}{\multirow{2}{*}{\begin{tabular}{@{}l@{}} \scriptsize{\textbf{sun} spring / summer (8183 / 5739), \textbf{snow} (7975), \textbf{rain} (8127),}\\
\scriptsize{\textbf{overcast} spring / summer / winter (7377 / 8118 / 7860)}\end{tabular} }}\\
\noalign{\vskip 4mm}
\hline
\noalign{\vskip 0.5mm}
\multirow{3}{*}{\scriptsize{\textbf{Test}}} & \multicolumn{1}{l}{\multirow{2}{*}{\begin{tabular}{@{}l@{}} \scriptsize{\textbf{sun} spring / summer / autumn (2406 / 1827 / 2416),}\\
\scriptsize{\textbf{overcast} spring / summer / winter (2178 / 2461 / 7407),}\\
\scriptsize{\textbf{dawn} (4851), \textbf{snow} (2450), \textbf{rain} (1949)}\end{tabular} }}\\
\noalign{\vskip 6.6mm}
\hline
\noalign{\vskip 0.5mm}
\multirow{5}{*}{\begin{tabular}{@{}l@{}}
\scriptsize{\textbf{spring}}\\
\scriptsize{\textbf{summer}}\\
\scriptsize{\textbf{autumn}}\\
\scriptsize{\textbf{winter}}\\
\\
\end{tabular}} & 
\multicolumn{1}{l}{\multirow{5}{*}{\begin{tabular}{@{}l@{}} \scriptsize{2015-03-10-14-18-10 / 2015-05-19-14-06-38,}\\ 
\scriptsize{2014-07-14-14-49-50 / 2015-08-13-16-02-58,}\\
\scriptsize{2014-11-18-13-20-12,}\\
\scriptsize{2014-12-09-13-21-02 / 2014-12-12-10-45-15 / 2014-12-02-15-30-08 /}\\ 
\scriptsize{2015-02-03-08-45-10 \change{/} 2015-02-13-09-16-26}\end{tabular} }}\\
\noalign{\vskip 12.5mm}
\bottomrule
\end{tabular}
\label{tab:dataset}
\vspace{-5mm}
\end{table}

\subsection{Dataset}
While \change{there exist} several datasets for visual place recognition in challenging conditions, most of them provide \change{only} {\it single, isolated images} as queries. By contrast, our approach requires {\it sequences of images} as queries in order to reconstruct the scene. Therefore, for training and evaluation, we use  sequences from the popular and challenging Oxford RobotCar Dataset~\cite{Maddern16ijrr}.
Within this dataset, the same \SI{10}{km} route through central Oxford was captured approximately twice a week over more than a year.
We choose a set of ten sequences that \change{represent} the large variance in visual appearance to be expected in a long-term navigation scenario, see Table \ref{tab:dataset}.
With approximately two sequences of the main route selected per season, our subset exhibits a large diversity in illumination, weather conditions including snow, sun, and rain as well as structural changes.
Our selection effectively compresses the about 100 traversals of the original dataset while preserving its challenging characteristics for visual place recognition.
We follow the common procedure of splitting each traversal into \textit{geographically non-overlapping} training, validation, and testing segments, resulting in approximately \SI{51}{K} training samples selected from seven sequences, \SI{17}{K} validation, and \SI{24}{K} testing samples chosen across ten sequences. %
\change{Images} are cropped to remove the hood of the car, and downscaled by a factor of two \change{when used as} visual CNN input (but not \change{when used as DSO input}).
Furthermore, fully overexposed images and long sequences during which the car is stationary are discarded to ensure tracking stability of DSO. %
The discretized volume around each camera pose is fixed at $40 \times 40 \times \SI{20}{m}$ with a grid resolution of $96 \times 96 \times 48$ voxels.

\subsection{Evaluation Methodology}
Our experiments evaluate place recognition based on pairwise matching across the sequences in our testing set.
All $45$ unique sequence combinations are taken into account.
Each model is evaluated in terms of exhaustive pairwise matching and nearest-neighbor retrieval.
To evaluate a model on a given sequence pair, we loop over the images of the first sequence while the images of the second sequence are used to build the database.
The best-performing model is selected based on the results obtained on the validation split while the testing set is used exclusively to obtain the final results.
\vspace{1mm}

\noindent \textbf{Exhaustive pairwise matching.}
Given a sequence pair, we evaluate how well a model discriminates between matching and non-matching descriptor pairs within the set of \textit{all} possible pairs where one descriptor comes from the query sequence and the other from the database sequence.
Each descriptor pair is classified into \textit{should} and \textit{should-not} matches: they should be matched if they represent locations with relative ground truth distance of less than \SI{5}{m} and heading difference of less than \SI{30}{degrees}.
In contrast, pairs with relative ground truth distance larger than \SI{20}{m} between the associated locations should not be matched.
Descriptor pairs corresponding to locations with relative distance between \SI{5}{m} and \SI{20}{m} and any relative heading can but do not have to be matched.
A descriptor pair is deemed to match if their $L_1$ distance is lower than a predefined threshold $d_{emb, th}$.
Consequently, matched descriptors are categorized into true and false positives while not matched descriptors are categorized as either true or false negatives.
This allows for computation of precision-recall (PR) curves parameterized by $d_{emb, th}$.
These are summarized using mean average precision (mAP), which is equivalent to the area under the PR curve.
\vspace{1mm}

\noindent \textbf{Recall@1 -- Nearest-neighbor retrieval.}
To verify the utility of the various descriptors, we follow the common procedure \cite{Arandjelovic16cvpr, Sattler12bmvc} to evaluate retrieval by looking at the $N$ nearest neighbors among all database descriptors for a given query descriptor.
It is deemed correctly recognized if there is at least one descriptor within the $N$ retrieved ones with associated ground truth distance below \SI{20}{m}.
For each sequence pair, we iterate over all descriptors of the query sequence and compute recall@N as the percentage of correctly recognized query descriptors.
Since some of the recorded sequences show deviations from the main route, we only consider query descriptors for which at least one truly matching database descriptor exists.
Due to place constraints, we restrict our analysis to the most difficult setting, $N=1$.

\setlength{\heavyrulewidth}{1.5pt}
\setlength{\abovetopsep}{6pt}
\begin{table}%
\footnotesize
\centering
\caption{Comparison of mean average precision (mAP) and recall@1 based on 128-d structural descriptors for varying depth $d_S$ of the feature extraction network and grid representations.}
\vspace{-4mm}
\begin{tabular}{c c c c c c c}
\toprule
\multirow{2}{*}{ \begin{tabular}{@{}c@{}} Depth $d_S$ \end{tabular} } & 
\multicolumn{3}{c}{mAP [-]} & \multicolumn{3}{c}{Recall@1 [\%]} \\
\noalign{\vskip 0.7mm}
& bo & so & ptc & bo & so & ptc\\
\noalign{\vskip 0.7mm}
\hline
\noalign{\vskip 1mm}
6 & 0.883 & -- & -- & 93.9 & -- & --\\
8 & 0.901 & -- & -- & \underline{94.5} & -- & --\\
9 & \underline{0.905} & 0.741 & 0.756 & \underline{94.5} & 90.4 & 89.5\\
10 & 0.879 & -- & -- & 93.1 & -- & --\\
12 & 0.867 & 0.759 & 0.731 & 92.3 & 90.5 & 89.3\\
\bottomrule
\end{tabular}
\label{tab:eval_grids}
\vspace{-6mm}
\end{table}
\setlength{\heavyrulewidth}{1.5pt}
\setlength{\abovetopsep}{6pt}
\begin{table*}%
\footnotesize
\centering
\caption{\change{Recall@1, in percent,} for different fusion methods against visual and structural descriptors, and several baseline methods, clearly showing performance gains when both input modalities are combined.
\changeAmadeus{See the supplementary material~\cite{Oertel20ralSupplementary} for a breakdown including each of the $45$ sequence pairings.}
}
\vspace{-4.5mm}
\begin{tabular}{c c c c c c c c c c}
\toprule
\noalign{\vskip 0.4mm}
\multicolumn{4}{c}{Composite descr. (ours)} & \multirow{2}{*}{ \begin{tabular}{@{}c@{}} Appearance \\ descr. \end{tabular} } & \multirow{2}{*}{ \begin{tabular}{@{}c@{}} Structure \\ descr. \end{tabular} } & \multirow{2}{*}{\begin{tabular}{@{}c@{}} NetVLAD \\ descr. (ft)\end{tabular} } &
\multirow{2}{*}{\begin{tabular}{@{}c@{}} \change{Multi-Process} \\ \change{Fusion} \end{tabular} } & \multirow{2}{*}{\begin{tabular}{@{}c@{}} DenseVLAD \\ descr.\end{tabular} } & \multirow{2}{*}{\begin{tabular}{@{}c@{}} SeqSLAM \\ \SI{40}{m} (\SI{20}{m}) \end{tabular} } \\
\cline{1-4}
\noalign{\vskip 0.7mm}
Concat & Weight. concat & Linear & MLP &  &  & & \\
\noalign{\vskip 0.5mm}
\hline
\noalign{\vskip 1mm}
\underline{98.0} & 96.7 & 96.9 & 95.7 & 94.2 & 93.9 & 90.0 & \change{89.3} & 83.1 & 73.1 (64.5) \\
\bottomrule
\end{tabular}
\label{tab:eval_fusion}
\vspace{-5mm}
\end{table*}
\section{Results and Discussion}
\label{sec:discussion}
\subsection{Voxel Discretization Method}
\label{sec:grid_type}
We separately train the structural feature extraction network for varying layer counts $d_S$ using the three grid representations described in Section \ref{sec:voxel_grids}.
To this end, the network learns by minimizing the margin-based loss given in \eqref{eq:margin_loss} for structural descriptors only.
Results averaged over all $45$ testing sequence pairs are reported in Table \ref{tab:eval_grids}.
We observe that using a binary occupancy representation achieves the best performance for all considered configurations.
\change{This is likely because the point count inside a voxel can vary depending on texture and appearance of the scene, which, unlike binary occupancy, is sensitive to seasonal changes.}
Hence, this representation is used in all other experiments.
\begin{figure}[htpb]%
	\vspace{-5mm}
	\centering
	\subfigure{%
		\includegraphics[width=0.49\linewidth]{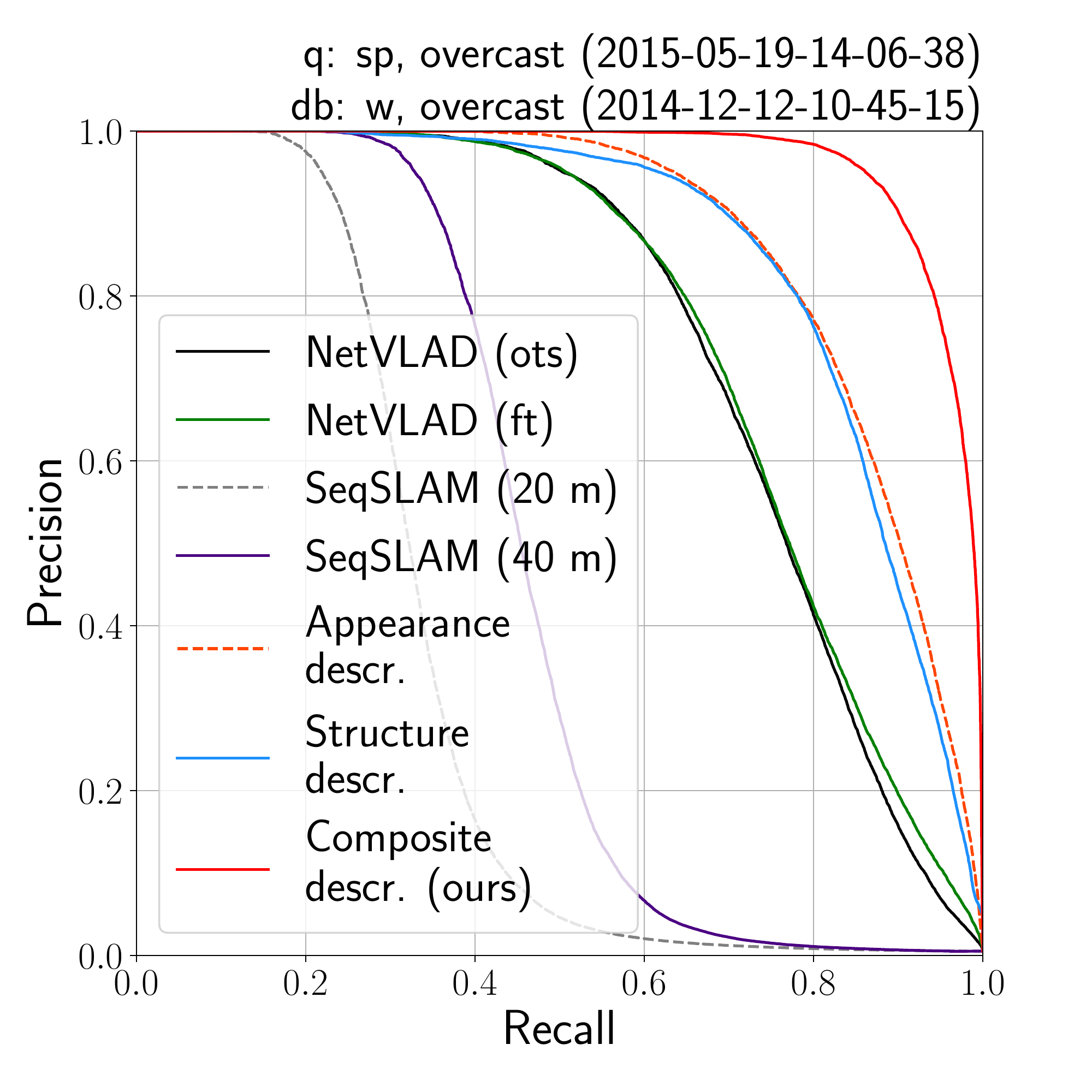}}%
	\hfill
	\subfigure{%
		\includegraphics[width=0.49\linewidth]{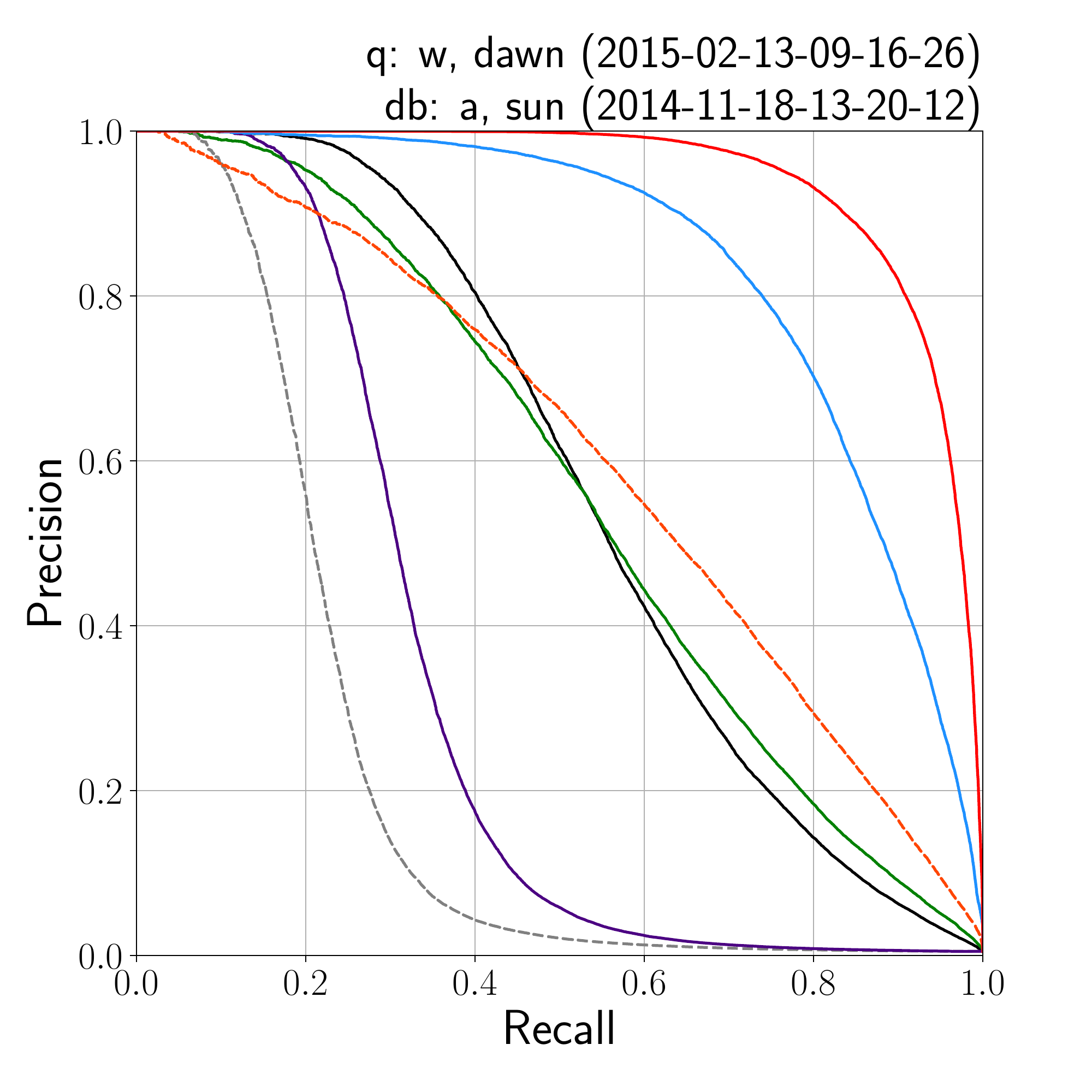}}%
	\vspace{-4.5mm}
	\caption{PR curves \changeAmadeus{from exhaustively matching two sequence pairs recorded in different seasons.}
	Generally, superiority of \changeAmadeus{using} either visual or structural descriptors varies \changeAmadeus{across sequence pairings}. 
	For all evaluated combinations, however, \changeAmadeus{fusing both visual and structural cues using our composite representation} results in significant performance gains.
	\change{See the supplementary material~\cite{Oertel20ralSupplementary} for the plots of all $45$ sequence combinations.}}
	\label{fig:pr_curves}
	\vspace{-4mm}
\end{figure}
\subsection{Feature Fusion}
\label{sec:feat_fusion}
To show the contribution of combining visual and structural features, we investigate the gains in retrieval using our composite descriptors over using only one of the two input modalities. 
Results are reported in Table \ref{tab:eval_fusion} and example queries and matches are show in Fig.~\ref{fig:retrieval_comparison}. 
We observe performance gains when using composite descriptors regardless of the fusion method used to generate them. 
In particular, we find that a simple concatenation of descriptors $g_A(I_i)$ and $g_S(G_i)$ results in the best overall performance with significant improvements over using only $g_A(I_i)$ or only $g_S(G_i)$.
\change{The fact that simple concatenation performs best could be because it forces both features to be learned -- other methods can degenerate into situations where parts of the input features are ignored.
Furthermore, more complicated fusion methods could be prone to overfitting, resulting in worse validation and testing performance.
Still, further investigation of feature fusion methods, including more advanced methods like attention refinement~\cite{Yu18eccv} are interesting future work.}
\changeAmadeus{Additionally, as exemplified in Fig. \ref{fig:pr_curves}}, we observe that for some of the sequence \changeAmadeus{pairings descriptors encoding structural cues perform better compared to those encoding visual features} while \changeAmadeus{for other pairings}, visual descriptors slightly outperform structural ones. 
\changeAmadeus{The left of Fig. \ref{fig:pr_curves} shows the PR curves resulting when exhaustively matching two sequences recorded during spring and winter, respectively. 
Both sequences are subject to similar illumination and weather conditions and we observe that our appearance- and structure-based descriptor variants perform very similarly. 
To the right of Fig. \ref{fig:pr_curves}, we provide the PR curves for a pairing again recorded during different seasons and additionally under drastically varying illumination and weather conditions. 
Images in the query sequence are subject to very low exposure while those in the database sequence are affected by direct sunlight resulting in greatly overexposed areas within the images. 
As a consequence, we observe the appearance-based variants to perform much worse compared to the structure-based descriptor. 
For both illustrated sequence pairings, our composite descriptor benefits from fusing both visual and structural cues into a unified representation.}
\changeAmadeus{We highlight that for \textit{each} of the 45 investigated sequence pairings}, our proposed composite descriptor consistently outperforms all other variants.
\vspace{-3.5mm}
\begin{figure}[htpb]%
	\centering
	\subfigure{%
		\includegraphics[width=0.49\linewidth]{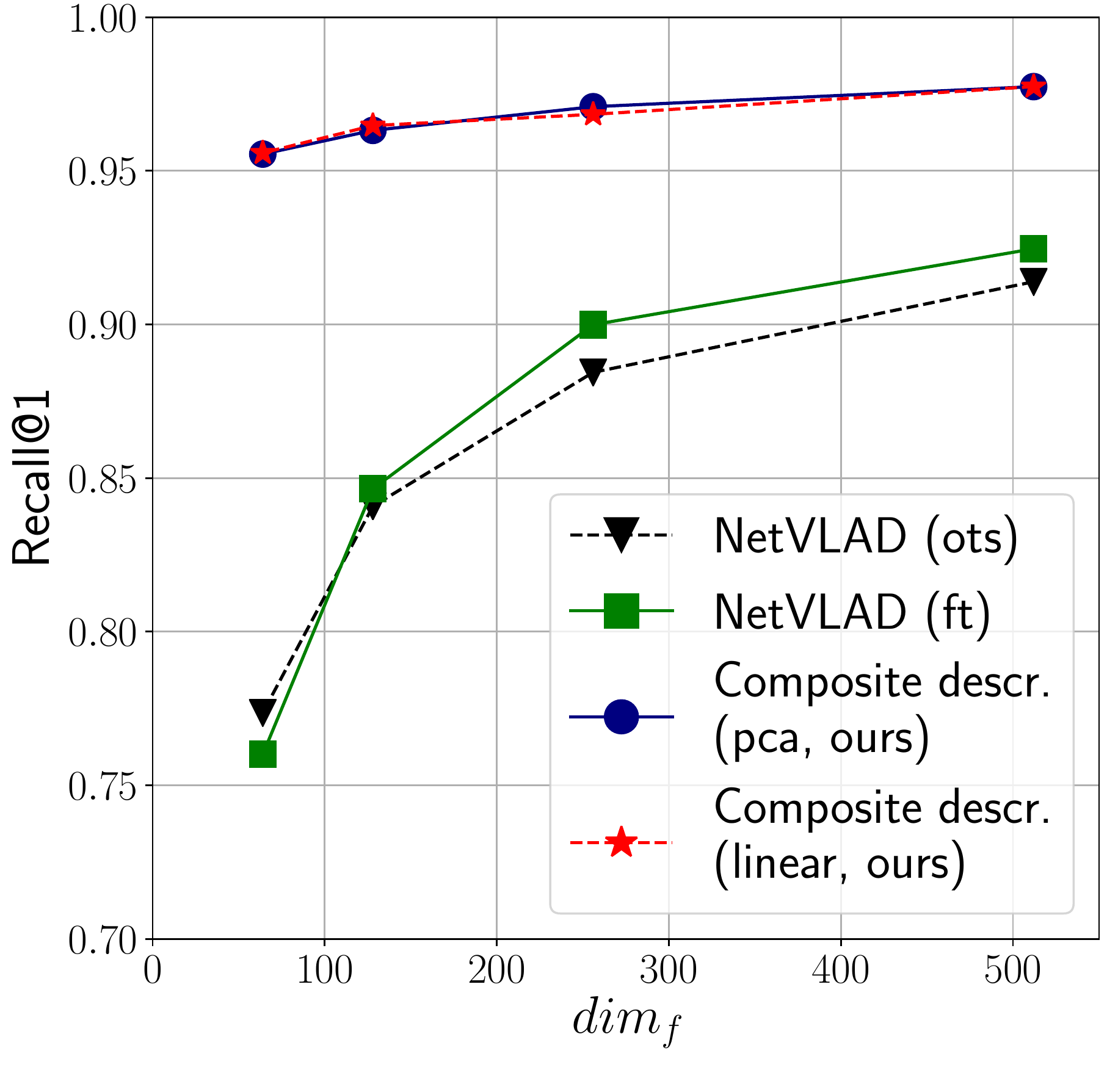}}%
	\hfill
	\subfigure{%
    		\includegraphics[width=0.49\linewidth]{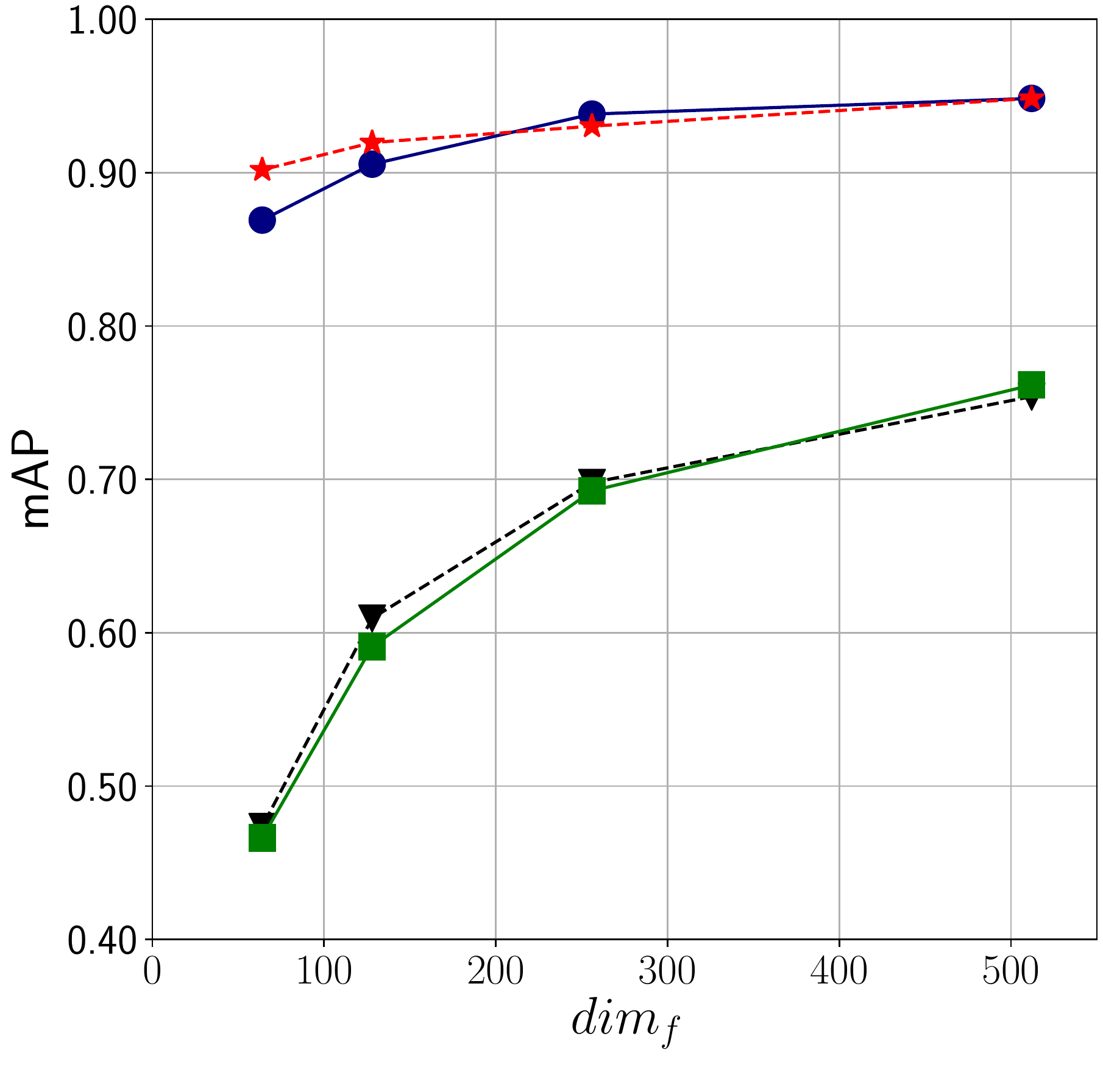}}%
	\vspace{-4.5mm}
	\caption{Mean average precision (mAP) and recall@1 shown for varying descriptor dimensionality $dim_f$. Both NetVLAD variants' performance degrades significantly more severely when reducing $dim_f$. This results in a relative gain of up to $90.5\%$ in mAP and 23.6\% in recall@1 of our composite descriptor against NetVLAD for $dim_f=64$.}
	\label{fig:dim_plot}
	\vspace{-5.5mm}
\end{figure}
\setlength{\heavyrulewidth}{1.5pt}
\setlength{\abovetopsep}{6pt}
\begin{table}[b]
\vspace{-4mm}
\footnotesize
\centering
\caption{\changeAmadeus{Average timings for computing a single instance of each descriptor variant measured using an Nvidia Titan Xp. Values in parentheses indicate computation times when descriptors are computed in batches.}}
\vspace{-3mm}
\begin{tabular}{p{0.15 \textwidth} p{0.2 \textwidth}}
\toprule
\textbf{Descriptor type} & \multicolumn{1}{l}{\textbf{Avg. computational cost per descriptor}}\\
\noalign{\vskip 0.5mm}
\hline
\noalign{\vskip 0.5mm}
Appearance & \multicolumn{1}{c}{\SI{8.96}{ms} (\SI{7.53}{ms})}\\
Structure & \multicolumn{1}{c}{\SI{11.93}{ms} (\SI{10.62}{ms})}\\
Composite, concat & \multicolumn{1}{c}{\SI{18.80}{ms} (\SI{17.50}{ms})}\\
NetVLAD & \multicolumn{1}{c}{\SI{60.82}{ms} (\SI{46.68}{ms})}\\
DenseVLAD & \multicolumn{1}{c}{\SI{1690}{ms} ( - )}\\
\bottomrule
\end{tabular}
\label{tab:timings}
\end{table}
\subsection{Baseline Comparison}
\label{sec:baseline}
We evaluate our \change{method} against SeqSLAM \cite{Milford12icra}, DenseVLAD \cite{Torii18pami}, NetVLAD \cite{Arandjelovic16cvpr} \change{and Multi-Process Fusion~\cite{Hausler19ral}}. 
\change{
We ensure that images are spaced by $0.5m$ to avoid problems when the car stands still.
We evaluate SeqSLAM sequence lengths of $40$ and $80$ frames (around $20m$ and $40m$) and use a linear trajectory search velocity of $0.8$ to $1.2$.
For DenseVLAD, we train the visual vocabulary on $25$ million RootSIFT descriptors extracted from the training set.
The final descriptors are projected to $256$ dimensions to match the dimensions of all other methods.
For the evaluation of Multi-Process Fusion, we use the publicly available VGG-16 trained on Places365 and keep the suggested default parameters.
The comparison in retrieval performance is given in Table \ref{tab:eval_fusion}. The reported values result from averaging the Recall@1 measured for each of the 45 sequence pairs.}
Note that \change{while this would suggest that our appearance-only branch outperforms NetVLAD,} these are results specific to the narrow-baseline Oxford Robotcar Dataset.
On the wide-baseline VGG Oxford Buildings dataset~\cite{Philbin07cvpr}, we have found for example that NetVLAD performs better than our appearance-only branch, with an mAP of $0.54$ versus $0.15$.
Unfortunately, our training data is restricted to narrow baselines, as we have not found a dataset that provides both image sequences and a wide variety of wide baseline matches.
As NetVLAD outperforms all of the other baselines, we thoroughly compare our composite descriptor to two variants of NetVLAD descriptors while also considering different descriptor dimensions $dim_f$.
Using few descriptor dimensions can significantly boost nearest neighbour search performance and reduce memory requirements.
The first version uses the publicly available \textit{VGG-16 + NetVLAD layer + PCA whitening} off-the-shelf weights trained on Pitts30k.
The second version represents the off-the-shelf weights after fine-tuning them on our training set for 15 epochs lasting approximately 70 hours on an Nvidia RTX 2080 Ti.
Following \cite{Arandjelovic16cvpr}, we use PCA to generate NetVLAD descriptors of varying dimensionality ranging between 64 and 512.
Even though concatenation of $g_A(I_i)$ and $g_S(G_i)$ results in best performance, training a linear projection allows us to precisely control the number of target dimensions $dim_f$ of our composite descriptor.
Hence, we individually train both feature extraction networks modified to produce 256- instead of 128-dimensional  descriptors.
We then initialize our compound network using these pretrained extraction networks and continue training different weight sets by varying $dim_f$.
By using the concatenation of $g_A(I_i)$ and $g_S(G_i)$, we obtain 512-dimensional composite descriptors. Additionally, we train three more variants each using a linear projection to control $dim_f$.
The results are illustrated in Fig. \ref{fig:dim_plot}, which shows that matching and retrieval performance of both NetVLAD variants severely diminishes when reducing descriptor dimensionality.
For $dim_f=256$ -- the best-performing compact projection dimension reported in \cite{Arandjelovic16cvpr} -- our composite descriptor outperforms NetVLAD by 39.1\% in exhaustive pairwise matching and 8.9\% in nearest-neighbor retrieval.
By decreasing $dim_f$, our approach outperforms NetVLAD by as much as 90.5\% and 23.6\% in matching and retrieval, respectively.
\change{We also evaluate the projection of our descriptor using a PCA trained on the same data as the NetVLAD PCA instead of a linear projection using our training method.
As seen in Figure~\ref{fig:dim_plot}, recall@1 is the same, while mean average precision is slightly worse.
Note, however, that the original dimension of NetVLAD ($4096$) is much higher than that of our descriptor ($512$).}
\vspace{-1.5mm}
\changeAmadeus{\subsection{Computational Costs}}
\changeAmadeus{Training the visual and structural feature extraction networks described in Section \ref{sec:net_details} using a single GPU requires approximately \SI{35}{hrs} and \SI{77}{hrs}, respectively. 
The compound architecture is initialized with the resulting weights and further trained for approximately \SI{27}{hrs}.
Note that a large fraction of training time is spent on batch sampling using the hard mining strategy detailed in Section \ref{sec:train_method}. Furthermore, the time spent on hard mining increases significantly as training progresses since such hard sample pairs become increasingly difficult to find.}
\changeAmadeus{In Table \ref{tab:timings}, we further provide an overview of average inference times, including those generated by the purely descriptor-based baselines NetVLAD and DenseVLAD.}

\section{Conclusion}
In this paper, we have proposed to augment visual place recognition using structural cues. 
We have shown that a concatenation of feature vectors obtained from appearance and structure performs best among the evaluated fusion methods.
Our approach is completely vision-based and does not require additional sensors to extract structure.
In all of our experiments, our composite descriptors consistently outperform vision- and structure-only descriptors alike, as well as all baselines.
Specifically, when comparing our composite descriptor against NetVLAD, the relative performance gain, especially at low descriptor dimensions, can be as high as $90.5\%$ and $23.6\%$ in exhaustive pairwise matching and nearest-neighbor retrieval.
The good performance at low dimensions means that our approach is particularly well suited to fast, large-scale nearest neighbour retrieval.

\addtolength{\textheight}{-0cm}   %

{\small
\bibliographystyle{ieeetr}
\bibliography{all}
}

\end{document}